\pdfoutput=1

\documentclass[11pt]{article}

\PassOptionsToPackage{dvipsnames,svgnames}{xcolor} %
\usepackage[preprint]{acl}

\usepackage{times}
\usepackage{latexsym}

\usepackage[T1]{fontenc}
\usepackage{amsfonts}
\usepackage[utf8]{inputenc}

\usepackage{microtype}

\usepackage{inconsolata}

\usepackage{graphicx}
\usepackage{booktabs}
\usepackage{multirow}
\usepackage{fancybox}

\usepackage{comment}
\usepackage{adjustbox}
\usepackage{listings}
\usepackage{fancyvrb}
\usepackage{fvextra}

\usepackage{amsmath}
\usepackage{cuted}
\usepackage{cleveref}
\usepackage{array}
\newcolumntype{C}[1]{>{\centering\arraybackslash}p{#1}}

\usepackage[dvipsnames,svgnames]{xcolor}
\definecolor{darkblue}{RGB}{0,0,150}
\definecolor{darkred}{RGB}{150,0,0}
\definecolor{darkgreen}{RGB}{0,150,0}
\definecolor{mybgcolor1}{HTML}{006d77}
\definecolor{mybgcolor2}{HTML}{edf6f9}

\usepackage{xspace}
\usepackage{subcaption}
\usepackage{makecell}

\newcommand{\predbench}{{\tt PredictaBoard}\xspace}

\title{\predbench: Benchmarking LLM Score %
Predictability}

\author{
  \textbf{Lorenzo Pacchiardi\textsuperscript{1}},
  \textbf{Konstantinos Voudouris\textsuperscript{1,2}},
  \textbf{Ben Slater\textsuperscript{1}},
  \textbf{Fernando Mart\'inez-Plumed\textsuperscript{3}},
\\
  \textbf{Jos\'e Hern\'andez-Orallo\textsuperscript{1,3}},
  \textbf{Lexin Zhou\textsuperscript{1,3}},
  \textbf{Wout Schellaert\textsuperscript{3}}
\\
\\
  \textsuperscript{1}Leverhulme Centre for the Future of Intelligence, University of Cambridge, United Kingdom\\
  \textsuperscript{2}Institute for Human-Centered AI, Helmholtz Zentrum Munich, Germany\\
  \textsuperscript{3}VRAIN, Universitat Polit\`ecnica de Val\`encia, Spain
\\
  \small{
    \textbf{Correspondence:} \href{mailto:lp666@cam.ac.uk}{lp666@cam.ac.uk}
  }
}

\everypar{\looseness=-1}  %

\begin{document}
\maketitle
\begin{abstract}

Despite possessing impressive skills, Large Language Models (LLMs) often fail unpredictably, demonstrating inconsistent success in even basic common sense reasoning tasks. This unpredictability poses a significant challenge to ensuring their safe deployment, as identifying and operating within a reliable ``safe zone'' is essential for mitigating risks. %
To address this, we present \textit{\predbench}, a novel collaborative benchmarking framework designed to evaluate the ability of score predictors (referred to as \textit{assessors}) to anticipate LLM errors on specific task instances (i.e., prompts) from existing datasets.
\predbench evaluates \textit{pairs} of LLMs and assessors by considering the rejection rate at different tolerance errors. 
As such, \predbench stimulates research into developing better assessors and making LLMs more predictable, not only with a higher average performance. We conduct illustrative experiments using baseline assessors and state-of-the-art LLMs. 
\predbench highlights the critical need to evaluate predictability alongside performance, paving the way for safer AI systems where errors are not only minimised but also anticipated and effectively mitigated. Code for our benchmark can be found at \url{https://github.com/Kinds-of-Intelligence-CFI/PredictaBoard}
\end{abstract}

\section{Introduction}

\begin{figure}[t!]
    \centering
    \includegraphics[width=1\columnwidth]
    {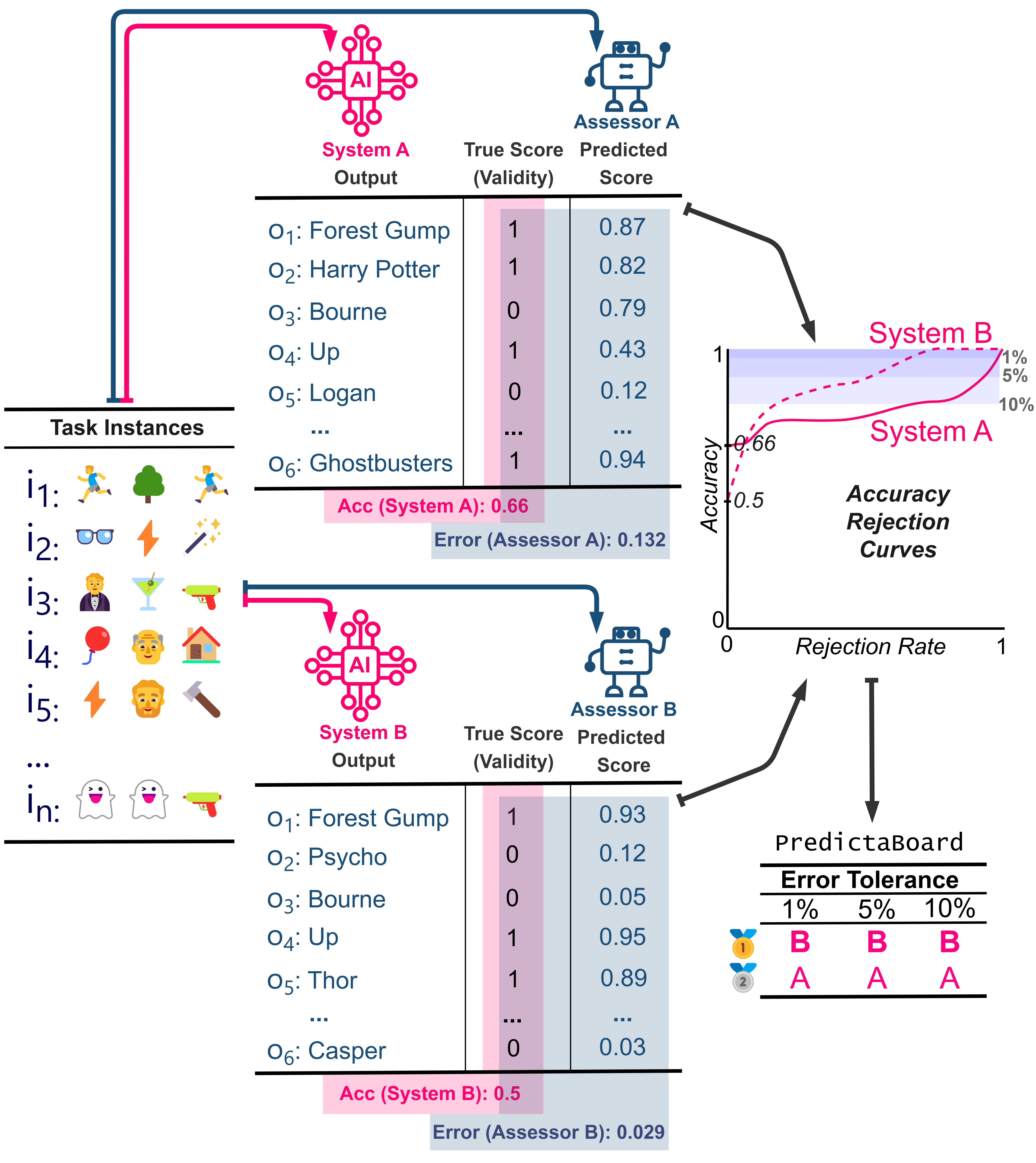}
    \caption{\predbench evaluates pairs of AI systems and score predictors (assessors). Top: System A is applied to a dataset of instances
    giving an accuracy of 0.66 (average of the scores). Assessor A attempts to predict the probability of success for each instance, with a mean squared error (Brier score) in these score predictions of 0.132. Bottom: System B is applied to the same dataset, giving an accuracy of 0.5. Assessor B for System B has an error in the score predictions of 0.029. While System A is better than System B on average, System A is much less predictable with Assessor A than System B with Assessor B. Considering the non-rejection rate at different tolerance errors (e.g., 1\%, 5\%  and 10\%), the pair $\langle$ System B, Assessor B $\rangle$ wins.}
    \label{fig:main}
\end{figure}

\looseness=-1
A key component of safety in high-stakes scenarios is  knowing the operating conditions of the system in use, namely, the specific \textit{task instances} in which the system succeeds \cite{leveson2002system, bahr2014system, hendrickx2024machine, zhou2024predictableartificialintelligence}. %
Consider, for example, two autonomous driving systems. System A always detects pedestrians correctly in day time but has a predictably high failure rate at night, allowing for safety measures (e.g., requiring a human intervention). System B, on the other hand, although generally better performing, has random, unpredictable failures that leave the driver uncertain about when to intervene. Even if System B has higher performance across the typical range of conditions encountered by drivers, System A is safer due to its predictability. This is visualised in a Q\&A domain in Figure \ref{fig:main}. %

\looseness=-1
This principle extends to frontier AI systems, such as Large Language Models\footnote{Many of which can also interpret images and, increasingly, audio and video.} (LLMs), which are rapidly being integrated into high-stakes scenarios \citep{kim2024language,huang2024benchmarking,javaid2024large}. Their growing adoption introduces the potential for high-consequence harm -- significant negative impacts on individuals, organisations, or society as a whole-- thus requiring robust safeguards to ensure both high performance and safety \cite{hendrycks2023overviewcatastrophicairisks}. Here as well, a key component to risk mitigation is {\em validity} predictability, namely, enabling users to anticipate and reject inputs that cause the AI systems to produce undesirable outputs or outcomes \cite{zhou2024predictableartificialintelligence, hendrickx2024machine, vafa2024largelanguagemodelsperform}; Sec~\ref{sec:predictability_safety} discusses this more in detail.

\looseness=-1
However, 
humans struggle to predict when LLMs will be correct \citep{carlini_gpt4_challenge}, can be biased by prior interactions \citep{vafa2024largelanguagemodelsperform} and even struggle to evaluate explanations provided by LLMs \citep{steyvers2025large}. 
Additionally, alignment using \textit{Reinforcement Learning from Human Feedback} (RLHF, \citealp{ouyang2022training}) and other techniques lead to LLMs potentially deceiving humans more often \citep{wen2024languagemodelslearnmislead, williams2024targeted}, %
resulting in unpredictable and unsafe behaviour %
\citep{anwar2024foundationalchallengesassuringalignment, zhou2024larger}.  

\begin{figure}[tb]
    \centering
    \includegraphics[width=\columnwidth]
    {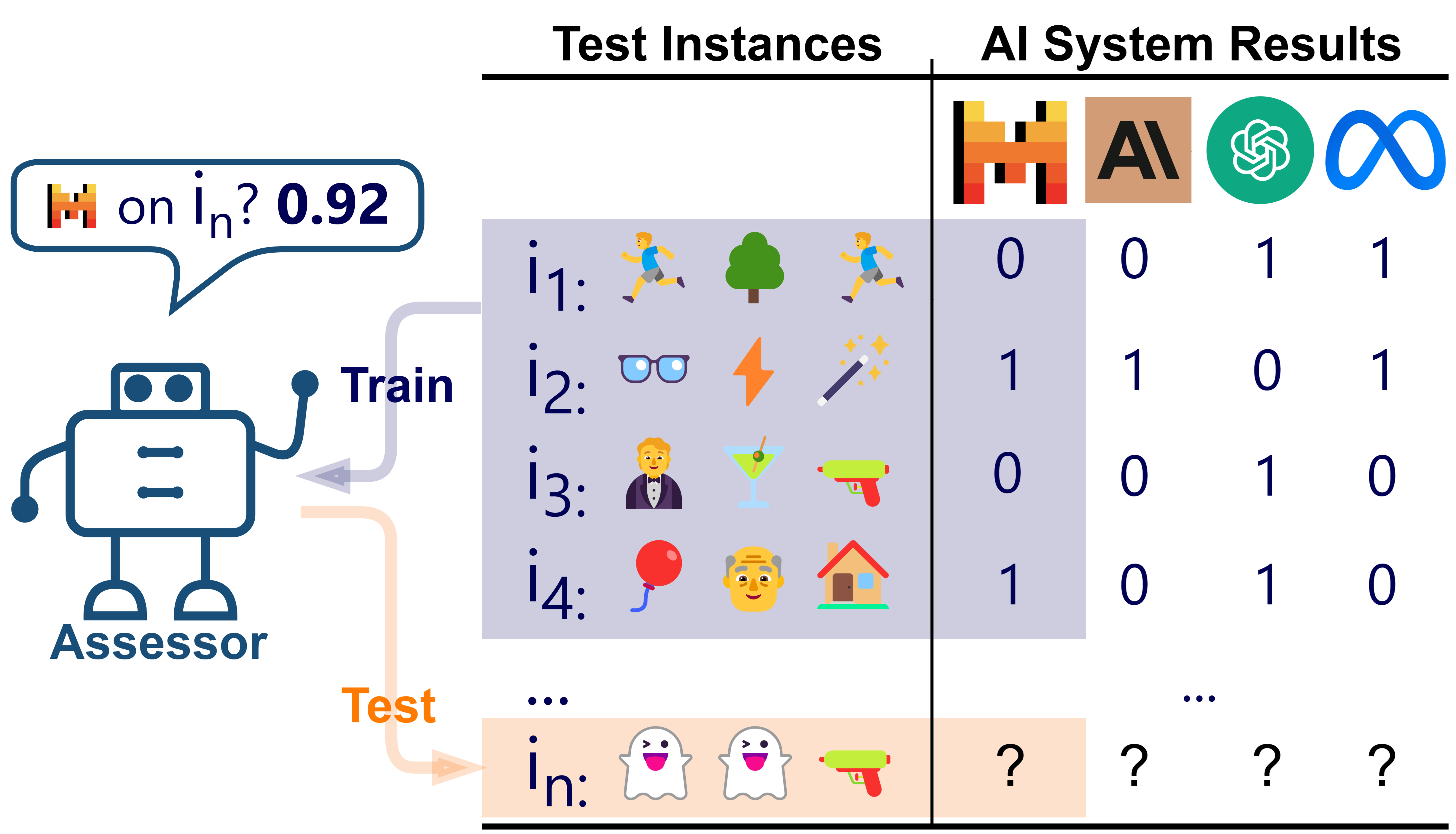}
    \caption{%
    Example of the score prediction problem. The displayed assessor is trained on the performance of a LLM on a split of the LLM test data, %
     allowing it to anticipate the LLM's failures on novel test instances. Assessors can however be built in different ways.}
    \label{fig:assessors}
\end{figure}

Technical solutions, including Uncertainty Quantification (UQ, \citealp{shorinwa}) or external \textit{assessors}
\citep{hernandez2022training} have been proposed to predict the success of and LLM on individual cases %
(see Figure~\ref{fig:assessors}). However, UQ methods require the input to be passed though the LLM, are not always reliable \citep{pawitan2024confidence, kapoor} and are degraded by RLHF \cite{tian-etal-2023-just}; on the other hand, assessors' performance is limited by LLMs' idiosyncrasies, such as 
prompt perturbation leading to wildly different results \cite{dhole2022nlaugmenterframeworktasksensitivenatural,shen2024donowcharacterizingevaluating} and success on difficult task instances not guaranteeing success on easy ones even within the same domain \cite{zhou2024larger}.
Most importantly, no standardised framework exists to track performance of validity prediction methods and thus stimulate research into better ones and more predictable LLMs.

\looseness=-1
To address this, 
we introduce \textit{\predbench}, the first collaborative benchmarking framework  that jointly assesses LLM performance and  the predictability of that performance. %
The subjects of \predbench are LLM-assessor pairs, 
with the latter predicting the LLM's score on each benchmark instance (i.e., prompt). 
Solely ranking pairs based on assessor's performance would however lead to the LLM that always fails (whose score is perfectly predictable) dominating. 
Hence, as metric, \predbench combines the LLM performance and the quality of the score predictions into an \textit{Accuracy-Rejection Curve} (ARC, \citealp{nadeem10ARCs}) representing the LLM's performance on instances for which the assessor predicts probability of success above different threshold values. 
From this curve, practitioners can determine the threshold for their error tolerance and determine the size of the \textit{predictably valid}\footnote{By `predictably valid', we refer to cases where the validity on a task instance can be reliably anticipated \citep{zhou2024predictableartificialintelligence}.} operating region (see Fig~\ref{fig:main}). Alternatively, the Area Under the ARC can be used  to combine performance of LLM-assessor pairs over all error tolerance levels  (Sec.~\ref{subsec:metrics}).

To compete in \predbench, we require assessors to not rely on the LLM's outputs. While these could be used without comparing them to ground truths, avoiding them entirely eliminates implicit reliance on the latter, makes assessors robust to
manipulation by the LLMs (similarly to how relying on human feedback makes LLM outputs more persuasive \citealp{wen2024languagemodelslearnmislead,williams2024targeted})
 and avoids wasted computations on instances where the LLM will predictably fail. However, we allow the use of internal LLM activations to the input. Essentially, this requirement distinguishes assessors from filters or verifiers of LLM answers and makes \predbench representative of the real world, where the ground truth is unknown.

To increase the predictably valid region, researchers can either 1) develop assessors that better identify patterns in task instances influencing LLM success, 2) train LLMs which are intrinsically more predictable, or 3) develop LLM-assessor pairs with joint high performance and predictability. To enable research in the first direction, \predbench provides instance-level results for state-of-the-art LLMs on popular datasets, allowing to test both in-distribution and out-of-distribution predictability (Sec.~\ref{subsec:dataset}). Alongside this, \predbench includes baseline \textit{anticipative} assessor architectures (predicting success before the LLM has seen the instance \citealp{hernandez2022training}), thus facilitating research in the second direction (Sec.~\ref{subsec:baseline_assessors}). A leaderboard ranking pairs of LLMs and assessors can be set up, thus ensuring overall progress in predictability. This paper reports initial results with the baseline assessors on state-of-the-art LLMs included in \predbench (Sec.~\ref{sec:experimental_results}).

The current version of \predbench focuses on %
success scores, %
but the same framework can be applied to safety and alignment benchmarks \citep{zhang-etal-2024-safetybench, mazeika2024harmbench}.
Moreover, the \predbench framework can be applied to any other type of AI system, such as LLM agents, embodied agents and vision systems. %
Ultimately, the long-term vision of \predbench is to shift AI benchmarking to consider predictability alongside performance, aligning it with traditional risk assessment practices that evaluate both factors \cite{leveson2016engineering,aven2016risk,amodei2016concrete}.

\section{AI Predictability and Safety}
\label{sec:predictability_safety}

In this paper, %
we focus on instance-level 
predictability of validity (see App.~\ref{app:predictability_def} 
for a formal definition). For LLMs, an instance is a specific input prompt, and validity can refer to any performance indicator (e.g. success or safety scores, or other metrics for biased or unethical outcomes). %
As shown in Figure \ref{fig:main}, for each LLM we build one or more \textit{assessors}\footnote{Also known as  ``rejector'' \citep{hendrickx2024machine}.} \citep{hernandez2022training} to predict the validity of the LLM output on a specific instance.

Validity predictability contributes to safety %
in high-stakes environments, filtering  out the inputs that lead to unacceptable behaviour can prevent harms---inputs can be rejected, redirected to a more reliable system, or supervised by humans. Thus, reliable assessors provide a cost-efficient safety layer of safety that complements built-in model mitigations  and post-generation filtering \cite{hernandez2022training, zhou2024predictableartificialintelligence}.

\looseness=-1
In terms of developing safer models, 
scalable oversight methods \cite{Leike2018ScalableAA, Christiano2018SupervisingSL, burns2023weak, kenton2024scalable} already use weaker AI models for high-quality feedback for training and overseeing complex AI models. Here, assessors can signal inputs where supervision may fail, or flag prompts that induce convincing but invalid outputs (e.g., in RLHF settings). %

Finally, while red-teaming efforts \cite{council2023ai}  are directed at finding harmful inputs, the ability to predict such inputs enables scalable vulnerability detection and design guardrails. %
While this could also be used by bad actors to exploit system vulnerabilities, the ability of safety researchers to more effectively address those vulnerabilities before deployment likely leads to a positive net effect in reducing risks. On the negative side, pairing AI systems with assessors may lead to over-reliance by conflating increased and absolute safety (automation bias). While this should be considered in system design and mitigated through user education, the safety benefits outlined above outweigh this concern.

\section{Related Work}
\label{sec:related_works}

\paragraph{Aggregate LLM performance prediction} Previous studies explored aggregate performance prediction across computational scales (\textit{scaling laws}, \citealp{kaplan2020scaling, hernandez2020measuring}) and predicted  LLMs' accuracy on BIG-Bench \citep{srivastava2023imitationgamequantifyingextrapolating} tasks using factors such as parameters or compute usage \citep{ye2023how,owen2024predictable}. Relatedly, \citet{ruan2024observational} %
predicted aggregate task performance using latent factors derived from benchmark performance and compute usage of multiple LLMs. 
In contrast, \predbench focuses on instance-level predictability.

\paragraph{How human users predict LLM performance}
Humans were found 
to only marginally beat random guess in predicting GPT-4's performance \citep{carlini_gpt4_challenge}. 
Relatedly, \citet{vafa2024largelanguagemodelsperform} showed humans overestimate LLM future performance based on prior interactions, especially with larger models in high-stakes contexts. They argue that ``the best LLM is the one that allows humans to make the most reliable inferences about where it will succeed'', closely aligning with \predbench's motivation. \citet{zhou2024larger} indicated that human predictions become unreliable as AI systems become more capable 
and \citet{steyvers2025large} found LLM-produced explanations supporting a statement do not lead humans to reliably assess whether that statement is correct, even when the LLM's token-level probabilities are calibrated. %
Finally, in a specific use case, \citet{bansalchallenges} reported that the failures of an LLM software engineer agent cannot be reliably anticipated.

\paragraph{Instance-level LLM performance prediction}
Various disciplines offer approaches for instance-level performance prediction. For instance, \citet{drapal2024MetaLearningNoveltyDetection} combined novelty detection with meta-learning to reject instances that are likely to cause %
AI failure. Additionally, Item Response Theory (IRT), originally developed to predict human performance %
\cite{embretson2013item}, has been adapted for machine learning and NLP \cite{martinez2019item,lalor2016building,kipnis2024metabench,polo2024tinybenchmarks,vania-etal-2021-comparing}, although it requires previously processed instances, limiting predictability for new inputs. %

Some works trained ``assessors'' to predict instance-level LLM performance. %
\citet{kadavath2022language} trained LLMs to predict the probability of succeeding on a question without reference to %
a specific answer, %
which performed satisfactorily but struggled with novel tasks. 
\citet{schellaert2024analysing} effectively predicts the performance of LLMs on 100+ BIG-bench tasks, outperforming subject systems in confidence, and maintaining predictability across different model sizes, suggesting scalability.
\citet{zhou2022reject}  showed that smaller LLMs can %
predict performance of %
larger models %
on certain tasks, almost halving errors and computational costs. %
\citet{DRAPAL2024112351} obtained explainable meta-rules from trained assessors to identify regions of predictable performance. 
Finally, while assessors allow to avoid prompting models on task instances causing failures, their training requires generating a failure/correctness dataset specific to each LLM. Single assessors  sharing information across many LLMs \citep{pacchiardi2024100instancesneedpredicting} reduce the instances on which each LLM must be tested.
All these techniques can be used as assessors for 
\predbench.

\paragraph{Factuality detection and correctness prediction with model internals}
Model internal activations have been used to detect statement truthfulness \citep{burns2022discovering, azaria-mitchell-2023-internal, Marks2023TheGO, burger2024truth}, deception \citep{goldowsky2025detecting} and hallucination \citep{ferrando2025do}. However, these works considered entire statements (or question–answer pairs), making them inherently non-anticipative. A few studies \citep{kadavath2022language, ferrando2025do} instead rely solely on the internal activations produced by the question to predict the correctness of the answer the model will generate; these can be viewed as assessors based on model internals, and can be evaluated within the \predbench{} framework, even though the baseline approaches we present here use only model-independent features of the input prompt. While tapping into the internal embedding may boost predictability, it hinges on (and thus exposes) the model’s own capacity to form confidence measures internally. In contrast, external assessors identify and exploit model-independent features that capture the intrinsic demands of the task.

\looseness=-1
\paragraph{Machine learning with reject options} %
Surveyed by \citet{hendrickx2024machine}, these models are closely related %
to the subjects of \predbench, with ``rejectors'' analogous to our interpretation of assessors. \citet{hendrickx2024machine} categorise rejectors according to their reliance on the ``predictor'' model.  %
Their ``separated'' rejectors are trained without involving the predictor, yet more powerful assessors %
can be trained using the observed LLM validity %
In any case, \citet{hendrickx2024machine} advocates for the %
development of benchmarks for %
ML with reject options, exemplified by %
\predbench, which can evaluate all kinds of rejectors and assessors.%

\paragraph{LLM Uncertainty Quantification (UQ)}
\looseness=-1
\citet{shorinwa} splits UQ methods for LLMs into token-level \citep{kadavath2022language}, verbalisation \citep{lin2022teaching, kapoor} and ``semantic similarity'' (prompting the model multiple times and grouping answers with the same meaning, \citealp{kuhn2023semantic}). 

However, in general, the performance of UQ methods is debated \citep{kapoor}; for instance, \citet{pawitan2024confidence} found  different methods to extract confidence to be poorly correlated and only partly indicative of correctness.
Additionally, in contrast to the anticipative assessors we use as baselines, %
UQ methods require inputs to be passed through the LLM, making them close to the ``integrated rejectors'' described in \cite{hendrickx2024machine}. Nevertheless, \predbench can be used to evaluate these methods too.

\paragraph{Reward models}
Reward modelling typically evaluates a model’s output against ``soft'' criteria such as toxicity \citep{faal2023reward} or alignment with user intent \citep{ziegler2019fine, ouyang2022training}; instead, the assessors subject of PredictaBoard predict the objective correctness of an answer, even though the validity notion PredictaBoard uses could be extended to include toxicity or related criteria. Additionally, PredictaBoard considers predicting the score from the question only, while reward modelling relies on the generated answer and is therefore more closely related to score modelling (automated grading, LLM-as-a-judge, \citealp{zhu2023judgelm}). Moreover, reward models are learnt independently of a particular LLM, while PredictaBoard’s assessors are specialized to anticipate a specific LLM’s behaviour.

\paragraph{LLM routing}
LLM routers \citep{lee2023orchestrallm,vsakota2024fly,lu2023routing,shnitzer2023LargeLanguageModel,ding2024hybrid}
direct task instances to the most appropriate LLM from a pool 
trading off performance, cost, response time or other factors\footnote{See \href{https://withmartian.com/}{Martian} for a commercially-available router.}.
This mechanism %
is similar to %
\textit{delegating classifiers}  where initial classifiers delegate difficult tasks to %
specialised ones 
\citep{ferri2004delegating}. Although assessors can act %
as routers by predicting LLM-specific probabilities of success, routers often bypass this step by directly selecting the model most likely to succeed. %
RouterBench \citep{hu2024routerbench} ranks routers based on their selection from a fixed set of LLMs, whereas \predbench evaluates each LLM paired with an  assessor.  %
While \predbench focuses on metrics %
measuring the size of operating conditions,  %
RouterBench uses an aggregate metric of %
quality and cost to optimise the use of %
LLMs in low-risk scenarios.

\looseness=-1
\paragraph{Model behaviour analysis} Various disciplines  %
help %
to understand the performance of %
AI models. %
 Surrogate modelling %
(\citealp{ilyas2022datamodels}) %
anticipates model behaviour from training data, while %
error analysis methods \cite{amershi2015modeltracker} %
identify %
weaknesses and %
incorrect predictions. Out-of-Distribution (OOD) detection \cite{hendrycks2016baseline,liang2017enhancing} targets %
unpredictable input behaviour (e.g., outliers). Unlike OOD methods, %
\predbench focuses on %
anticipating performance for inputs both in and out-distribution.

\paragraph{Guaranteed/SafeGuarded AI} Different research agendas \citep{dalrymple2024guaranteedsafeaiframework, darpa2024} propose the development of ``safeguarded'' AI systems, that come with (possibly probabilistic, \citealp{bengio2024bayesianoraclepreventharm}) performance guarantees in particular operating regions. \predbench can empirically test these methods.

\section{\predbench}
\label{sec:predbench}

\subsection{Dataset}
\label{subsec:dataset}

Our dataset consists of the instance-level performances of various LLMs on {\tt MMLU-Pro} \cite{wang2024mmluprorobustchallengingmultitask}  and the BIG-Bench-Hard (BBH, \citealp{suzgun2022challengingbigbenchtaskschainofthought}), for a total of 11383 and 5761 instances respectively. The results for 38 LLMs for both  MMLU-Pro and BBH were obtained from HuggingFace's \textit{Open LLM Leaderboard v2}, which ranks open-source LLMs on these benchmarks; further, the results for 3 versions of GPT-4o for MMLU-Pro were obtained from the original repository (Table \ref{tab:llms} in the Appendix includes the full list).
To ensure fair comparison, \predbench includes fixed randomly-sampled train, validation and test splits %
of MMLU-Pro: assessors can be trained and selected using the training and validation splits, and the performance on the test splits is reported. Additionally, we use the whole of BBH as Out-Of-Distribution (OOD) data to evaluate assessors trained on the train split of MMLU-Pro.

\subsection{Metrics}
\label{subsec:metrics}
While \predbench %
primarily employs metrics jointly assessing the validity of the LLM and the quality of the assessor's score predictions (\S\ref{subsubsec:combined_metrics}, to be used for the forthcoming associated competition), it also includes LLM-only metrics (\S\ref{subsubsec:subject_accuracy}) and assessor-only metrics (\S\ref{subsubsec:assessor_metrics}), as the best choice of metrics varies based on the considered application's requirements. %

Let $x_i\in \mathcal X$ denote some features of the $i$-th instance, with $\mathcal{X}$ denoting the space of instance features, and let $v_{i}\in\{0,1\}$ denote the validity (in our case, the success in providing the correct answer) of the considered LLM on instance $i$. %
An assessor $a: \mathcal{X} \to [0,1] $ estimates the probability $P(v_{i}=1|x_i)$. Assume we have $n$ instances on which the assessor is tested and the subject scored (i.e., a dataset \((x_i, v_{i})_{i=1}^n\))

\subsubsection{LLM-Only: Accuracy}
\label{subsubsec:subject_accuracy}
The \textit{accuracy} of the LLM is the average success over the dataset.

\subsubsection{Assessor-Only: Brier Score, AUROC}
\label{subsubsec:assessor_metrics}
The following assessor-only metrics treat the assessor as a probabilistic binary classifier.

\paragraph{Area Under the ROC Curve} (AUROC, \citealp{bradley1997use}), %
    evaluating an assessor's discrimination ability between positive and negative labels. 
    As the value of AUROC for perfect and random assessors are insensitive to label distribution, it can be seamlessly used to compare assessors for LLMs with different accuracy. Details in App.~\ref{app:metrics_auroc}. 
\paragraph{Brier Score} (BS, \citealp{gneiting2007strictly}) measures the mean squared error between the assessor predictions and the actual success:%
    \[
    \operatorname{BS} = 
    \frac{1}{n} \sum_{i=1}^n \left(a(x_i) - v_{i}\right)^2
    \]%
    A perfect assessor achieves a BS of 0, and larger scores indicate poorer predictions. 
    The BS can be decomposed into calibration and refinement  components (the latter is related to AUROC). However, its scale depends on the ratio of positive to negative labels, thus making it inconvenient to directly compare across LLMs. Details in App.~\ref{app:metrics_brier}. 
\paragraph{Winkler's Score} \citet{winkler} introduced a transformation of the BS which, in our case, relies on the average LLM success, thus making the score comparable across LLMs (formulation in App.~\ref{app:metrics_winkler}). The resulting score is maximised to 1 for a perfect assessor and 0 for an assessor predicting the average LLM success; negative values indicate worse performance than the average.

\subsubsection{Combined: ARC and PVR}
\label{subsubsec:combined_metrics}

\begin{figure}[!ht]
    \centering
    \includegraphics[width=\linewidth]{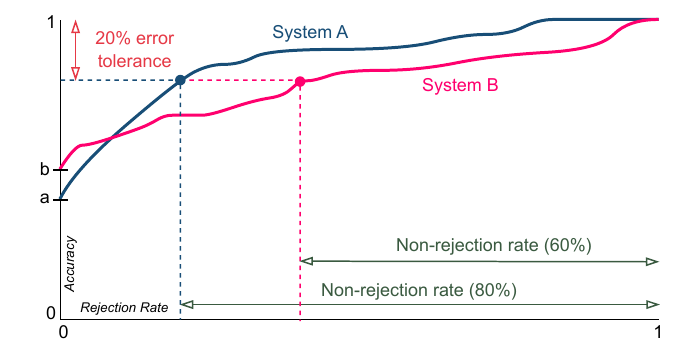}
    \caption{Accuracy Rejection Curves (ARC) allow graphical comparison of system accuracy based on rejection rates. At 20\% error tolerance, System A (blue) has a rejection rate of 80\% and System B (pink) has a rejection rate of 60\%. %
    }
    \label{fig:ARC}
\end{figure}

An Accuracy-Rejection Curve (ARC, \citealp{nadeem10ARCs}) is built by varying the rejection threshold of the assessor and computing the accuracy of the LLM on the non-rejected instances. The x-axis represents the rejection rate (0 to 1), while the y-axis shows the accuracy on non-rejected instances. The ARCs always converge at (1, 1), indicating 100\% accuracy at 100\% rejection rate. They start at (0, $acc$), where $acc$ is the accuracy without rejection. 
An example comparing two systems is shown in Figure~\ref{fig:ARC}.

The ARC shows the rejection/performance balance of the LLM-assessor pair. To obtain a single score for ranking, we look at the \textit{non-rejection rate} for a given error tolerance. This is the proportion of accepted instances at the lowest threshold (corresponding to the minimum rejection rate) that ensures  LLM errors are below the desired error level, such as 20\%. See Figure~\ref{fig:ARC} for details. To account for different error tolerances in different applications, we provide the non-rejection rates at 5\%, 10\% and 20\% error tolerance levels. We also refer to this metric as the Predictably Valid Region (PVR) up to a given error tolerance.

PVR is a suitable metric when safety is important and an error threshold can be established. To aggregate across all rejection thresholds, we suggest using the \textit{Area under the ARC} (AUARC), which is in $[0,1]$ and is maximised by a perfect LLM.

\subsection{Baseline Assessors}
\label{subsec:baseline_assessors}

\predbench includes several baseline \textit{anticipative} assessors \citep{hernandez2022training}, which predict the success of the LLM before it is exposed to inputs. These assessors go beyond simply avoiding reliance on the LLM's output; they also operate without access to internal activations, making their training architecture-agnostic. We encourage researchers to explore and develop assessors that leverage internal activations for potentially enhanced predictive capabilities.

In particular, we build assessors leveraging input text embeddings\footnote{We also fine-tuned small LLMs to predict the success of another LLM. As the performance of these was not competitive with the other assessors we do not report these, however those results are available in our repository.}. We plan to 
 incorporate additional approaches as they are developed (such as few-shot LLMs or extrapolating performance from similar examples) in future releases.

To obtain representations we  train assessors on, we used four different embedding schemes \citep{kusner2015word}: \textit{OpenAI} embeddings\footnote{Endpoint \texttt{text-embedding-3-large}.} \citep{neelakantan2022text} generated by models developed by OpenAI; \textit{Word2Vec} \citep{mikolov2013efficient} for learning word embeddings using neural networks; \textit{Fasttext} \citep{bojanowski2017enriching} that considers subword information; and \textit{n-grams} \citep{sidorov2014syntactic} using contiguous sequences of \textit{n} items.

Table \ref{tab:classifiers} shows the classifiers we trained to be our assessors, with each of the four  embeddings. %
Our 41 LLMs, 4 embedding schemes and 3 classifiers, gave us 492 LLM-assessor pairs in our baseline.

\begin{table}[h]
    \centering
    \caption{Classifiers used as assessors.}
    \resizebox{\columnwidth}{!}{%
    \begin{tabular}{@{}p{0.2\columnwidth}p{0.3\columnwidth}p{0.5\columnwidth}@{}} 
        \toprule
        \textbf{Id} & \textbf{Classifier} & \textbf{Hyperparameters}\\ \midrule
        \textbf{LR-l1}  & \makecell{Logistic \\ Regression}  & {\tt{solver= `liblinear', penalty = `l2'}} \\ 
        \textbf{LR-l2}  & \makecell{Logistic \\ Regression}  & {\tt{solver=`liblinear', penalty = `l1', C=1}}\\ 
        \textbf{XGB} & XGBoost &  {\tt{Default}} \\ \bottomrule
    \end{tabular}
    }
    \label{tab:classifiers}
\end{table}

\subsection{Testing New LLMs or Assessors}
\label{subsec:test_subjects_assessors}
By relying on the baseline assessors provided in \predbench, researchers can easily evaluate a new LLM. At the same time, researchers can develop novel assessor methods using \predbench's comprehensive collection of instance-level LLM results. This flexibility facilitates independent research into both areas. Additionally, entirely new LLM-assessor pairings can be evaluated.

\section{Experimental Results with Baselines and Existing LLMs}
\label{sec:experimental_results}

This section presents the results with our LLM-assessor baseline pairs. We trained assessors on the training split of the MMLU-Pro dataset and the scores of each LLM; then, we compute metrics on the test split of MMLU-Pro and on BBH. %

\subsection{In-Distribution Evaluation}
\begin{figure}[!ht]
    \centering
    \includegraphics[width=0.97\linewidth]{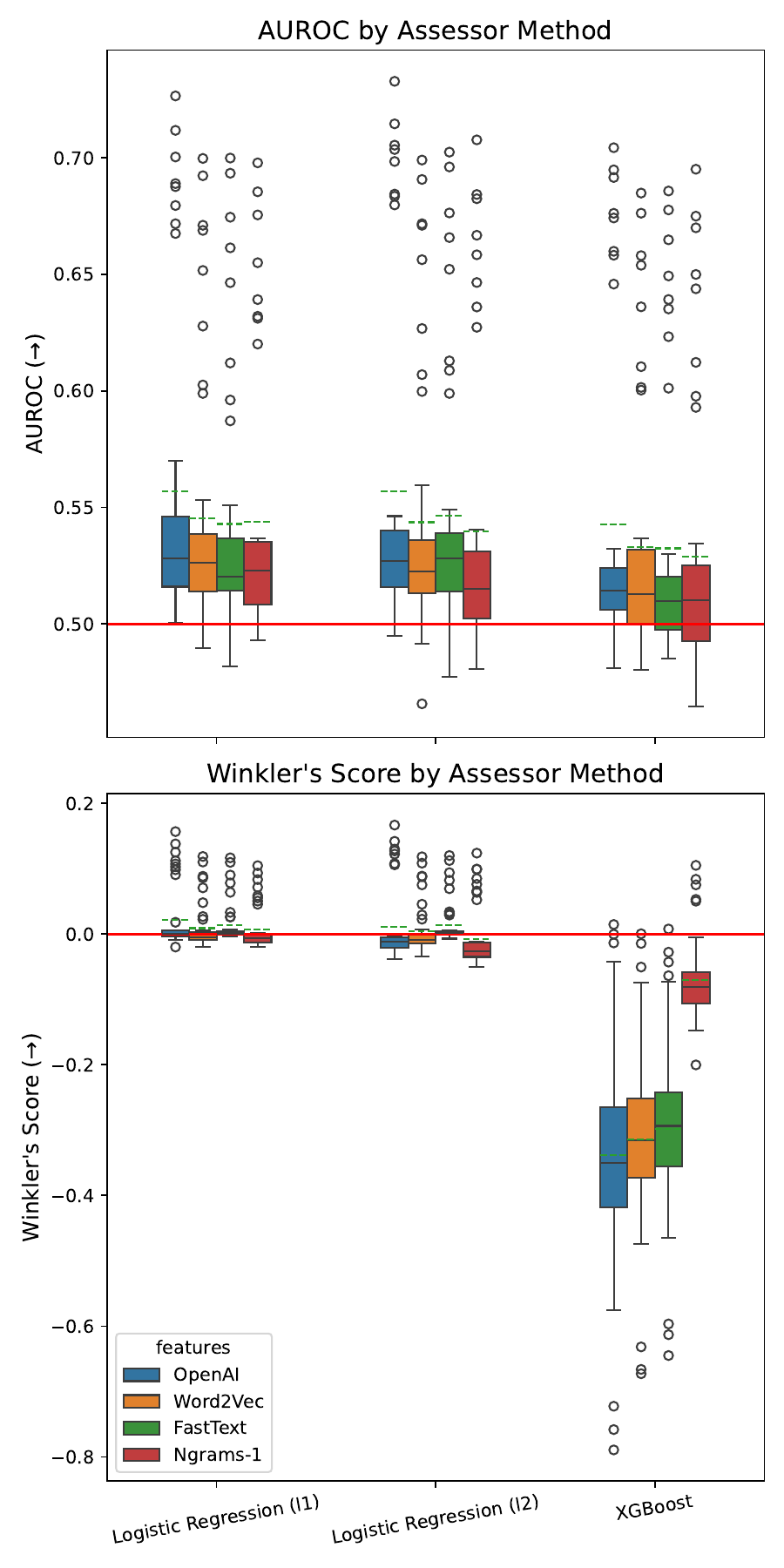}
    \caption{Distribution (over LLMs) of assessor-only performance metrics across different embedding schemes and classifiers. Top: AUROC; the red line shows the expected performance of a random assessor. Bottom: Winkler's score by classifier type; the red line shows the expected performance of a constant assessor predicting the LLM accuracy. Each boxplot displays median, quartiles, support and outliers of the distribution, while the green dashed line shows the mean.}
    \label{fig:experiments_AUROC_Brier}
\end{figure}

Firstly, we compare assessor methods by considering the distribution of assessor-only metrics over the LLMs. Figure \ref{fig:experiments_AUROC_Brier} shows the distribution of the AUROC and the Winkler's score. Most LLM-assessor pairs are slightly better or worse than random or constant baselines, with only a few being noticeably better (AUROC near 0.7 or Winkler's score near 0.15). No choice of embeddings method consistently outperforms the other, while the XGBoost classifier performs worse in terms of Winkler's score (indicating issues with calibration).

\begin{figure}[tb]
    \includegraphics[width=\linewidth]{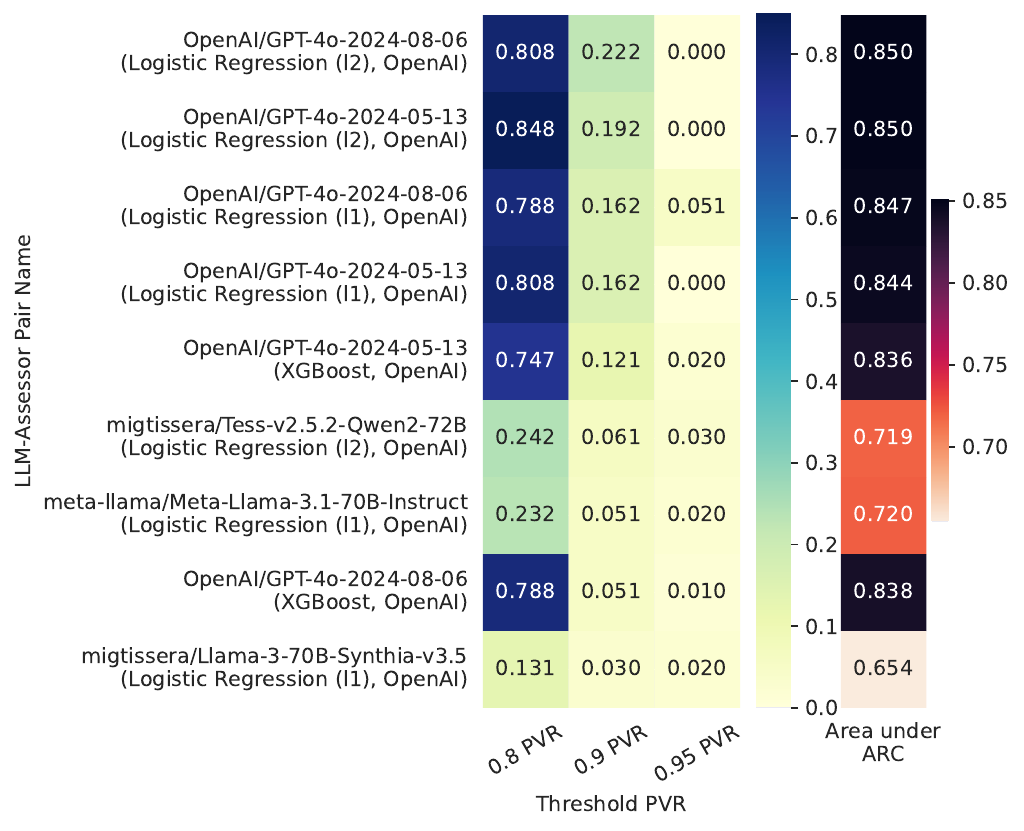}%
\caption{PVR at thresholds of 0.8, 0.9, and 0.95, and the area under the ARC curve for the union of the top 5 top LLM-assessor pairs at each threshold (in-distribution).}
        \label{fig:experiments_PPR}
\end{figure}

\looseness=-1
Next, to score LLM-assessor pairs, we consider the size of the PVR and the Area Under the ARC (AUARC). %
In Figure~\ref{fig:experiments_PPR} we show the size of PVR at thresholds 0.8, 0.9 and 0.95\footnote{At higher thresholds, PVR drops to near zero for all assessors. However, very low tolerance rates are crucial when AI systems pose catastrophic risks \citep{hendrycks2023overviewcatastrophicairisks}. In future iterations, we will include higher thresholds and encourage researchers to design AI systems and assessors capable of operating under minimal error tolerance.} for the 5 top subject-assessor pairs at each threshold; the AUARC is also reported to capture predicability across the full range of error thresholds. Notice how, as expected, the best pairs all include LLMs in the upper quartile in terms of average accuracy (see Figure~\ref{fig:llm_accuracy} in the Appendix). In particular,  LLM-assessor pairs get a good score at a threshold of 0.8. This is to be expected when the LLMs are fairly good at the task (the best LLM has 75\% accuracy), as the assessor can predict success most of the time. When the threshold is raised to 0.9 and above, we see a drastic drop in PVR, as this poses a greater requirement for assessors to make predictions that the LLM will fail. It is interesting, however, how the pair ranking is not preserved across the different thresholds.

\begin{figure}[!ht]
    \centering
    \includegraphics[width=\linewidth]{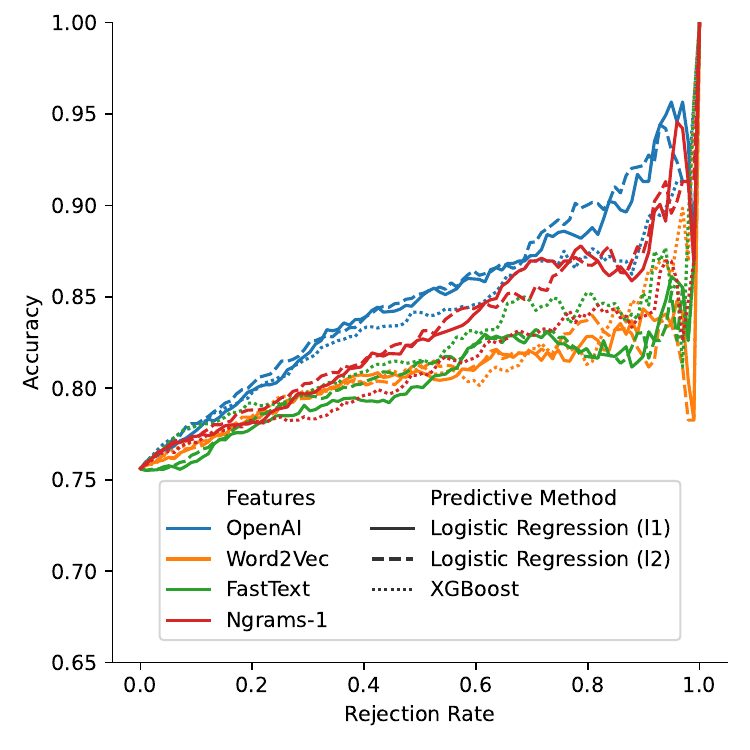}
    \caption{A comparison of the ARC curves for the different assessors of the highest accuracy LLM (``OpenAI/GPT-4o-2024-08-06'') in our dataset.}
    \label{fig:experiments_ARC_top_scorer}
\end{figure}

\begin{figure}[!ht]
    \centering
    \includegraphics[width=0.95\linewidth]{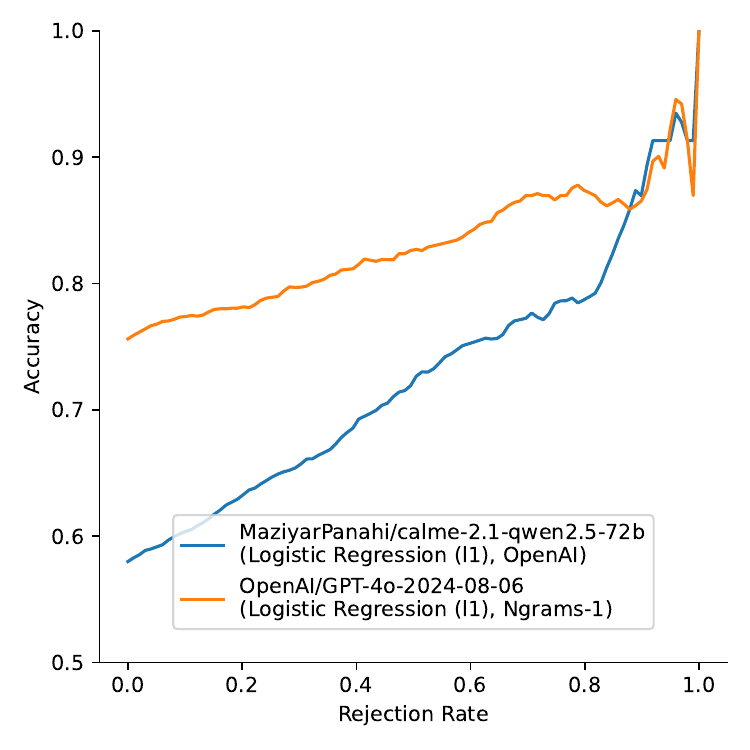}
    \caption{A comparison of two ARC curves in which one has a better performance at a PVR threshold of 0.8, and the other at a PVR threshold of 0.9.}
    \label{fig:experiments_ARC_crossing_pair}
\end{figure}

To demonstrate how the choice of assessor impacts the ARC, in Figure \ref{fig:experiments_ARC_top_scorer} we compare the ARCs obtained with different assessors for the LLM ``OpenAI/GPT-4o-2024-08-06'', which achieves the highest accuracy on MMLU-Pro. The ARC varies substantially depending on assessor. In Figure \ref{fig:experiments_ARC_crossing_pair}, we examine two selected examples from our baselines. These show a case in which one of the two LLM-assessor pairs has a better PVR at a threshold of 0.8, and the other at a threshold of 0.9.

\subsection{Out-of-Distribution Evaluation}

\begin{figure}[!ht]
    \includegraphics[width=\linewidth]{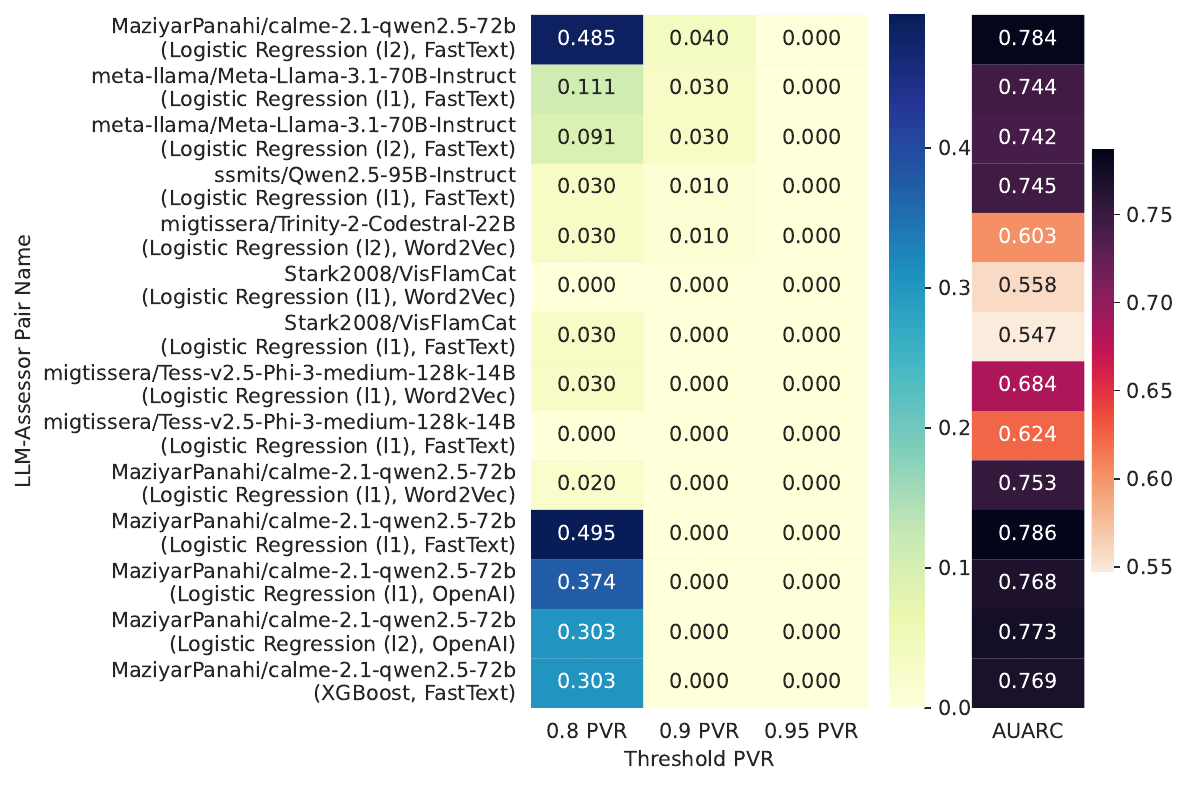}%
    \caption{PVR at thresholds of 0.8, 0.9, and 0.95, and the area under the ARC curve for the union of the top 5 top LLM-assessor pairs at each threshold (OOD).}
    \label{fig:experiments_PPR_OOD}
\end{figure}

To assess the robustness of our LLM-assessor pairs, we evaluated them on BBH after training on the train split of MMLU-Pro. Figure \ref{fig:experiments_PPR_OOD} replicates Figure \ref{fig:experiments_PPR} for the BBH benchmark. We use the same selection criteria as for the in-distribution results: the union of the top 5 PVR performers at each threshold. The lower values in Figure \ref{fig:experiments_PPR_OOD} highlight the difficulty of predicting performance OOD. Additionally, the top pairs differ from the in-distribution ones, suggesting the latter may not have the highest generalisation power. Other metrics for this OOD scenario are available in App.~\ref{app:ood_res}. %

\section{Conclusion}
\label{sec:conclusion}
\vspace{-0.1cm}
\looseness=-1
\predbench introduces a novel benchmark concept, evaluating pairs of models and validity predictors (\textit{assessors}). This aligns with the notion of validity predictability \citep{zhou2024predictableartificialintelligence}, highlighting how uncertainty on errors or safety, initially perceived as aleatoric, can become epistemic through pattern discovery \citep{hullermeier2021aleatoric}. %
Leveraging external predictors to extract instance-level patterns contributes to making AI systems predictable and understandable, something that intrinsic uncertainty estimation falls short of. This stresses the critical importance of joint progress on LLMs and their assessors. An LLM is only as predictable as the quality of its assessors, and an assessor method is only effective if it performs well for state-of-the-art LLMs. \predbench creates a unique opportunity to explore advancements in both LLMs and assessors, offering potential gains on these two fronts. %

\looseness=-1
This paper aimed to release an initial version of the benchmark and allow the community to guide its future extensions, that can  possibly include developing assessors that function across multiple LLMs and predicting safety indicators \citep{zhang-etal-2024-safetybench, mazeika2024harmbench}, rather than performance.
Additionally, comparing with human performance as assessors, as examined in recent studies \citep{carlini_gpt4_challenge,vafa2024largelanguagemodelsperform,gao2024take, zhou2024larger}, could provide insights into the differences in LLM predictability from automated and human perspectives.

\section{Limitations}\label{sec:limitations}

There are some limitations in the current version of \predbench. First, our baselines only include external anticipative assessors. In principle, \predbench allows for non-external assessors (including self-confidence and latent LLM's embeddings, as discussed in Sec.~\ref{sec:related_works}) and can be easily extended to consider the output of the model as well, using ``verifiers'' instead of assessors (although not relying on outputs has some advantages, highlighted in Sec.~\ref{sec:predictability_safety}). Additional conditions could also be considered, such as the invertibility of the assessor—-whether, by using the assessor, one can generate inputs that ensure the model's performance exceeds a given score, given constraints on the input. This capability would be particularly valuable for red-teaming applications or enhancing explainability (e.g., generating counterfactuals). Considering variations for these additional conditions and others (like the computational cost of the LLM-assessor pair) could 
allow to study properties of the assessors, such as the exploration of scaling laws for pairs of LLMs and assessors, or explore situations where the assessor cannot be  $n$ times more costly than the LLM. We leave all these considerations for future versions and specific competitions.

Some obvious limitations are the number of metrics, datasets and baseline methods. For metrics, one could also consider Pareto-dominance rather than single metrics, plotting the evolution of LLMs and assessors bidimensionally. Similarly, we could have considered cost-based metrics, such as those discussed in \citealp[Section 3.3]{hendrickx2024machine}, assigning a relative cost to rejections with respect to errors and computing the total cost using a fixed rejection threshold. 
Unlike the non-rejection rate metric we use (which is a specific case of cost-based metrics), these metrics require the definition of application-specific rejection costs and a maximum total cost, making them more complex to use in a standardised benchmark. For datasets, it is not always easy to find good sources covering a wide range of state-of-the-art LLMs at the instance level \cite{burnell2023rethink}, but more and more benchmarks with instance-level test results are available to be included, such as the remaining benchmarks involved in the Open-LLM Leaderboard, among others. These could be used to train assessors on more diverse data as well as for evaluating them out of distribution. In future releases, we thus plan to expand the coverage of our analysis and the datasets included in \predbench.

\newpage
\section*{Acknowledgments}
Lorenzo Pacchiardi received funding from US DARPA (grant HR00112120007, RECoG-AI) and Open Philanthropy, and computational support from OpenAI through the Researcher Access Program. Ben Slater received funding from an ESRC scholarship (ES/P000738/1). Konstantinos Voudouris received funding from the Templeton World Charity Foundation grant (Major Transitions in the Evolution of Cognition: TWCF-2020-20539) and was supported by Helmholtz Zentrum München Deutsches Forschungszentrum für Gesundheit und Umwelt (GmbH).
Jose Hernandez-Orallo and Fernando Martínez-Plumed received funding from CIPROM/2022/6 (FASSLOW) and IDIFEDER/2021/05 (CLUSTERIA) funded by Generalitat Valenciana, 
the EC H2020-EU grant agreement No. 952215 (TAILOR), 
and Spanish grant PID2021-122830OB-C42 (SFERA) funded by MCIN/AEI/10.13039/501100011033 and ``ERDF A way of making Europe" 
CÁTEDRA ENIA-UPV en IA DESARROLLO SOSTENIBLE, TSI-100930-2023-9.
This research project has benefitted from the Microsoft Accelerate Foundation Models Research (AFMR) grant program.
In compliance with the recommendations of  \citet{burnell2023rethink} on the reporting of evaluation results in AI, we include all the results at the instance level (\url{https://github.com/Kinds-of-Intelligence-CFI/PredictaBoard}).

\bibliography{custom}

\clearpage
\appendix

\section{AI ecosystems, predictability and assessor models}
\label{app:predictability_def}

In this appendix, we provide formal definitions of predictability as used in this paper, adapted from \citet{zhou2024predictableartificialintelligence}. In this regard, we model the AI system and its interactions within an \textbf{AI ecosystem} (ranging from single AI systems interacting with individual users for specific tasks to complex socio-technical environments, with different levels of granularity) as follows: $\mathcal{I}$ is the set of problem instances (e.g., input prompts). $\mathcal{S}$ is the set of AI systems considered (e.g., LLMs). $\mathcal{U}$ is the set of users or operators interacting with the AI systems. $\mathcal{O}$ is the set of possible outputs from the AI systems. $\mathcal{R} \subseteq \mathcal{I} \times \mathcal{S} \times \mathcal{U} \times \mathcal{O}$ capture the relationships and interactions among instances, systems, users, and outputs. An AI ecosystem at time $t$ is then represented as a tuple 
$
\mathsf{E}_t = \langle \mathcal{I}_t, \mathcal{S}_t, \mathcal{U}_t, \mathcal{R}_t \rangle
$
where the components may change over time.

We consider a distribution over ecosystems denoted as $\mathcal{E}_{t}$, and the complete history of interactions up to time $t$ is represented by:
$
\mathsf{H}_{\leq t} = \langle \mathsf{E}_{\leq t}, \mathsf{O}_{< t}, V_{< t} \rangle,
$
where $V_t$ is a random variable indicating the validity of outputs at time $t$ (e.g., whether the LLM provides a correct answer for instance $i$).

\textbf{Predictability} is then formalised through the conditional probability distribution:
$
p(V_{t+h} \mid \mathsf{H}_{\leq t}),
$
which represents the probability of observing a valid output at a future time $t+h$, given the history up to time $t$. We define unpredictability $\mathbb{Q}$ as the minimum expected loss over a family of predictors $\mathcal{F}_b$  constrained by resource budgets $b$:

\begin{equation}\label{eq:ecosystemunpred}
\mathbb{Q}(p,  {\cal H}_t,{\cal F}_b) := 
\min_{\hat{p} \in {\cal F}_b}
\underset{\substack{{\sf H}_{\leq t} \sim {\cal H}_{\leq t} \\ v \sim p(V_{t+h}|{\sf H}_t)}}{\mathbb{E}}  S (\hat{p}(V_{t+h}\:|\:H_{\leq t}), v) 
\end{equation}

\noindent where $S$ is a scoring function assessing the accuracy of predictions, such as the Brier Score.

In our context of \predbench, an \textbf{assessor model} $a$ belongs to a family of predictors $\mathcal{F}_b$, constrained by computational resources and the information it relies on (e.g., not using the LLM's outputs). An assessor is thus defined as $a: \mathcal{I} \rightarrow [0,1]$, predicting the validity of the LLM's output on individual instances, namely $P(v_i = 1 \mid x_i)$), where $x_i$ is the feature representation of instance $i$, and $v_i \in \{0,1\}$ indicates the validity (e.g. success) of the LLM on that instance. The goal is thus to find an assessor that minimises unpredictability $\mathbb{Q}$ by minimising the expected loss:
$
\min_{a \in \mathcal{F}_b} \frac{1}{n} \sum_{i=1}^n S\big(a(x_i), v_i\big),
$
over a test set of $n$ instances, where $S$ is the scoring function.

\section{Assessor metrics}
\label{app:metrics}

\subsection{Area Under the Receiving Operating Characteristic Curve (AUROC)}
\label{app:metrics_auroc}
The AUROC assesses the quality of a binary probabilistic classifier by measuring its ability to discriminate between positive and negative instances across various thresholds. Specifically, the AUC plots the True Positive Rate (TPR) against the False Positive Rate (FPR) at different threshold levels on this probability, obtaining a curve known as the Receiver Operating Characteristic Curve (ROC) curve. The AUROC is then calculated as the integral of this curve, providing a single scalar value that summarises the overall performance of the considered binary classifier. 
An AUROC of 1 indicates perfect discrimination, while an AUROC of 0.5 suggests no better performance than random chance. These extreme values are insensitive to the ratio of positively and negatively labelled instances in the dataset; thus, the AUROC can be seamlessly used to compare different scenarios where those ratios differ. This characteristic is particularly useful when comparing (LLM, assessor) pairs.
However, the AUROC is insensitive to monotonic transformations of the probabilities predicted by the classifier. This implies that a miscalibrated classifier can still achieve a high AUROC. While increasing the AUROC will enhance the discrimination between the two classes, it does not necessarily improve the calibration of the classifier.

\subsection{Brier Score}
\label{app:metrics_brier}
The Brier Score (BS) is equivalent to computing the mean squared error between the assessor predictions for each instance $x_i$ and the actual success $v_i$:

\[
\operatorname{BS} =  \frac{1}{n} \sum_{i=1}^n \left(a(x_i) - v_i\right)^2.
\]

A perfect assessor would achieve a BS of 0, and larger scores indicate poorer predictions. The BS is an example of a strictly proper scoring rule \cite{gneiting2007strictly} — that is, a scoring method for probabilistic predictions that encourages recovery of the true data distribution when minimised. As such, the BS can be decomposed into calibration and refinement components (with the latter related to AUROC). This decomposition means that the BS incentivises assessors to improve both calibration and discrimination.

However, the scale of the BS depends on the ratio of positive to negative elements $v_i$ for the considered subject. Specifically, an assessor that always predicts the proportion of positive samples $q$ in the dataset achieves a BS of $q(1 - q)$ as $n \to \infty$.%

\subsection{Winkler's score}
\label{app:metrics_winkler}
\citet{winkler} presented a generic way to correct binary scoring rules so that the score achieved by assessors that always predict the average success rate for subjects with different success rates is the same, and so can be easily compared. This relies on transforming a symmetric score (namely, for which $S(p,1)= S(1-p,0)$) into a non-symmetric one. Applying this transformation to the Brier Score leads to \cite[Sec. 3.2]{gneiting2007strictly}: 

\[
    \operatorname{WS} = 
    \frac{1}{n} \sum_{i=1}^n \frac{\alpha_i}{\beta_i},  
\]
where
\[
\begin{aligned}
    \alpha_i &= [(1-c)^2 - (1-a(x_i))^2 ]\mathbf{1}\{v_i=1\} \\&\quad + (c^2 - a(x_i)^2 )\mathbf{1}\{v_i=0\},\\
    \beta_i &= c^2 \mathbf{1}\{a(x_i)\le c\} + (1-c)^2 \mathbf{1}\{a(x_i)> c\},
\end{aligned}
\]
where $ c $ is the average accuracy of the considered LLM and $\mathbf{1}$ is the indicator function. 
 The score for the assessor predicting the observed success rate is 0 while that for a perfect assessor is 1.

\section{LLMs}

Our dataset consists of the instance-level performances of various LLMs on {\tt MMLU-Pro} \cite{wang2024mmluprorobustchallengingmultitask}  and the BIG-Bench-Hard (BBH, \citealp{suzgun2022challengingbigbenchtaskschainofthought}), for a total of 11383 and 5761 instances respectively. The results for a selection of 38 open-source LLMs for both  MMLU-Pro and BBH were obtained from HuggingFace's \textit{Open LLM Leaderboard v2}, which ranks open-source LLMs on these benchmarks; 
the selection includes a diversity of models from several well-known families, each with different architectures (LLama, GPT, Qwen, Mistral, etc.) and parameter sizes (up to 95B parameters). 
Further, the results for 3 versions of GPT-4o for MMLU-Pro were obtained from the original repository \cite{wang2024mmluprorobustchallengingmultitask}. Table \ref{tab:llms} includes the full list of considered LLMs.

\begin{figure*}[tb]
    \centering
    \includegraphics[width=\linewidth]{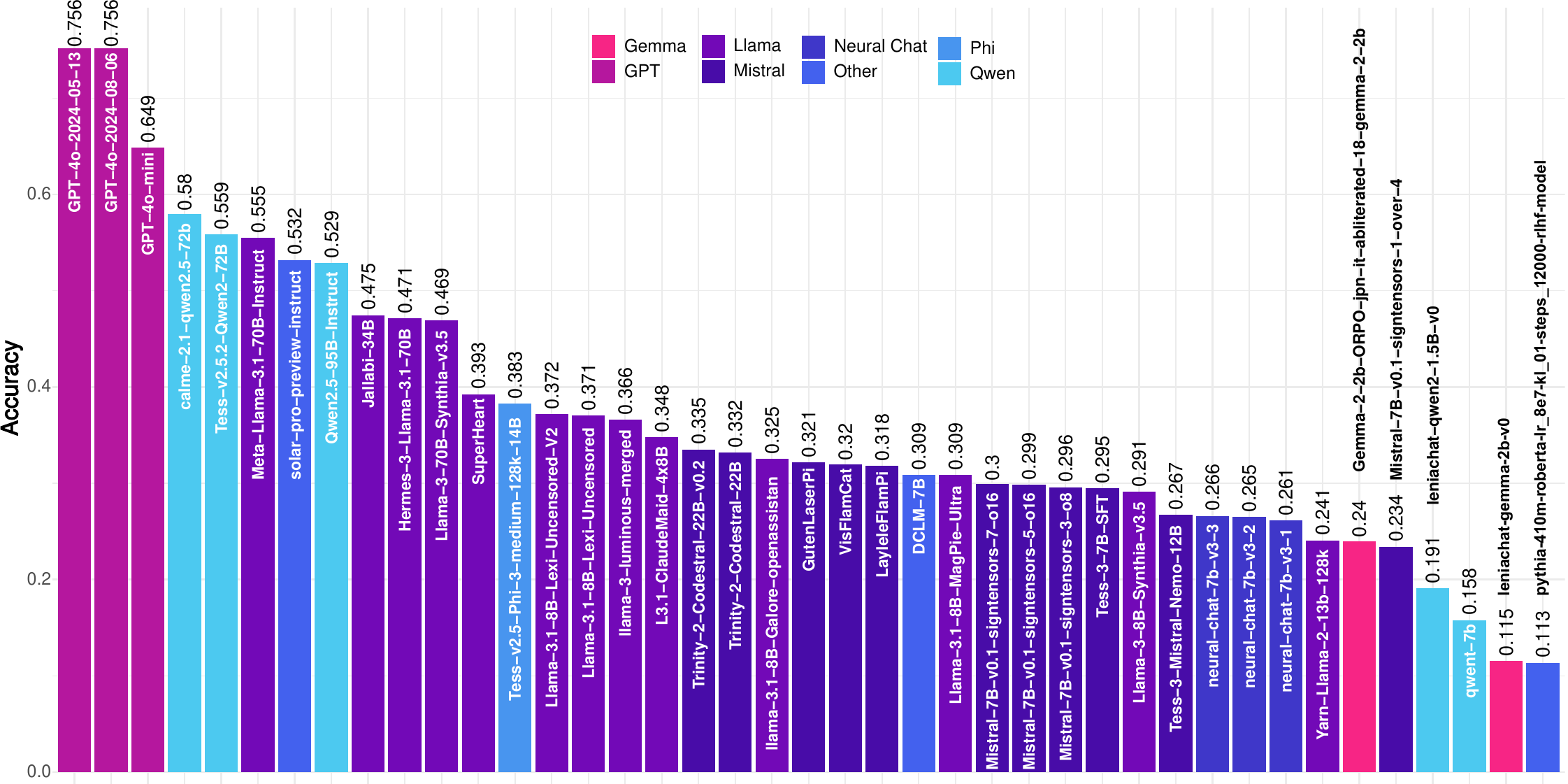}
    \caption{The performance of the LLMs in the MMLU-Pro dataset, expressed as the proportion of questions answered correct.}
    \label{fig:llm_accuracy}
\end{figure*}

\begin{figure*}[!ht]
    \centering
    \includegraphics[width=\linewidth]{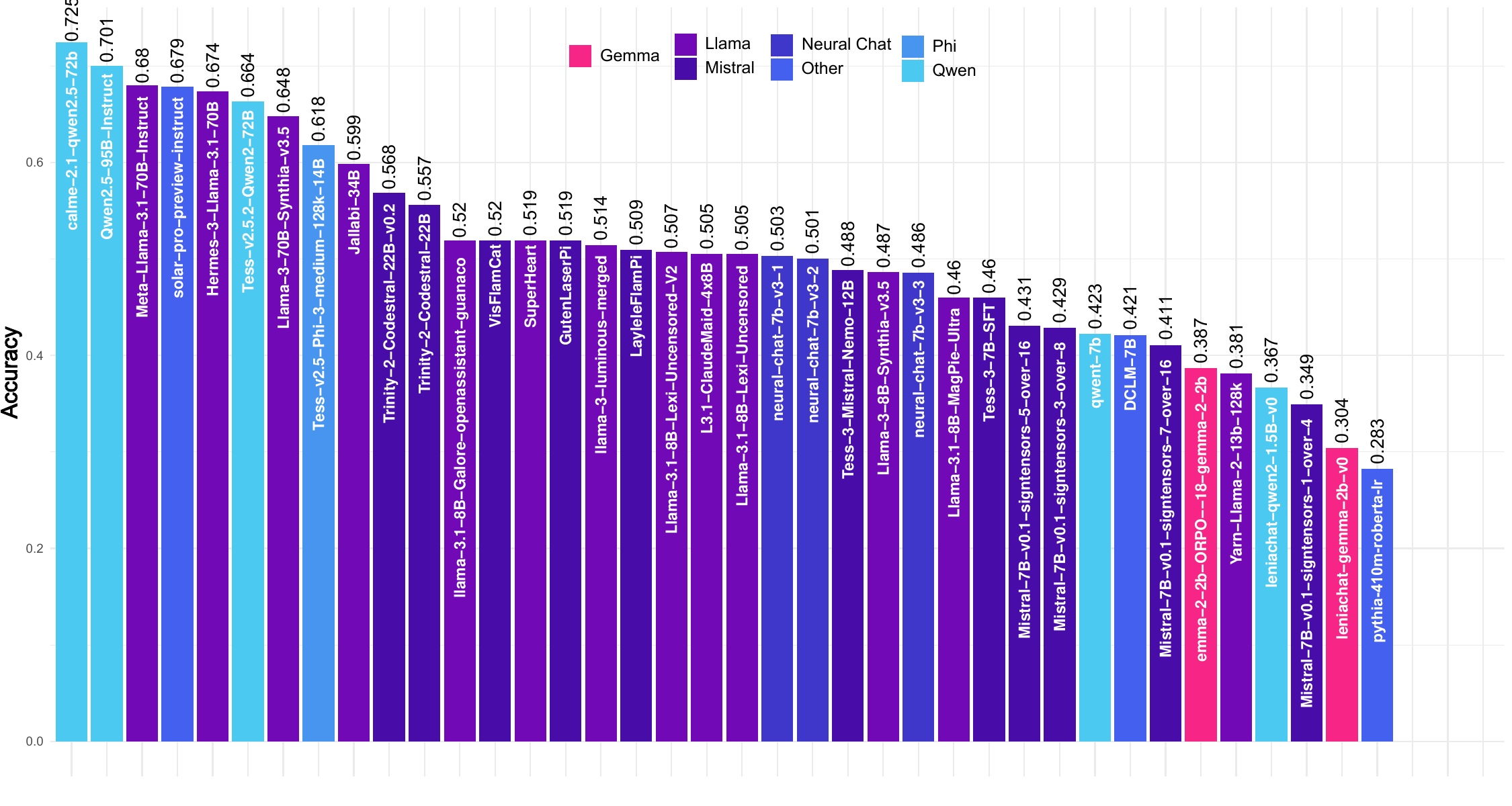}
    \caption{The performance of the LLMs in the BBH dataset, expressed as the proportion of questions answered correct.}
    \label{fig:llm_accuracy_ood}
\end{figure*}

Moreover, Figures \ref{fig:llm_accuracy} and \ref{fig:llm_accuracy_ood} show the performance of the different LLMs on the {\tt MMLU-Pro} and  BBH benchmarks respectively. 

\section{OOD AUROC and Winkler's score}
\label{app:ood_res}

\begin{figure}[tb]
    \centering
    \includegraphics[width=\linewidth]{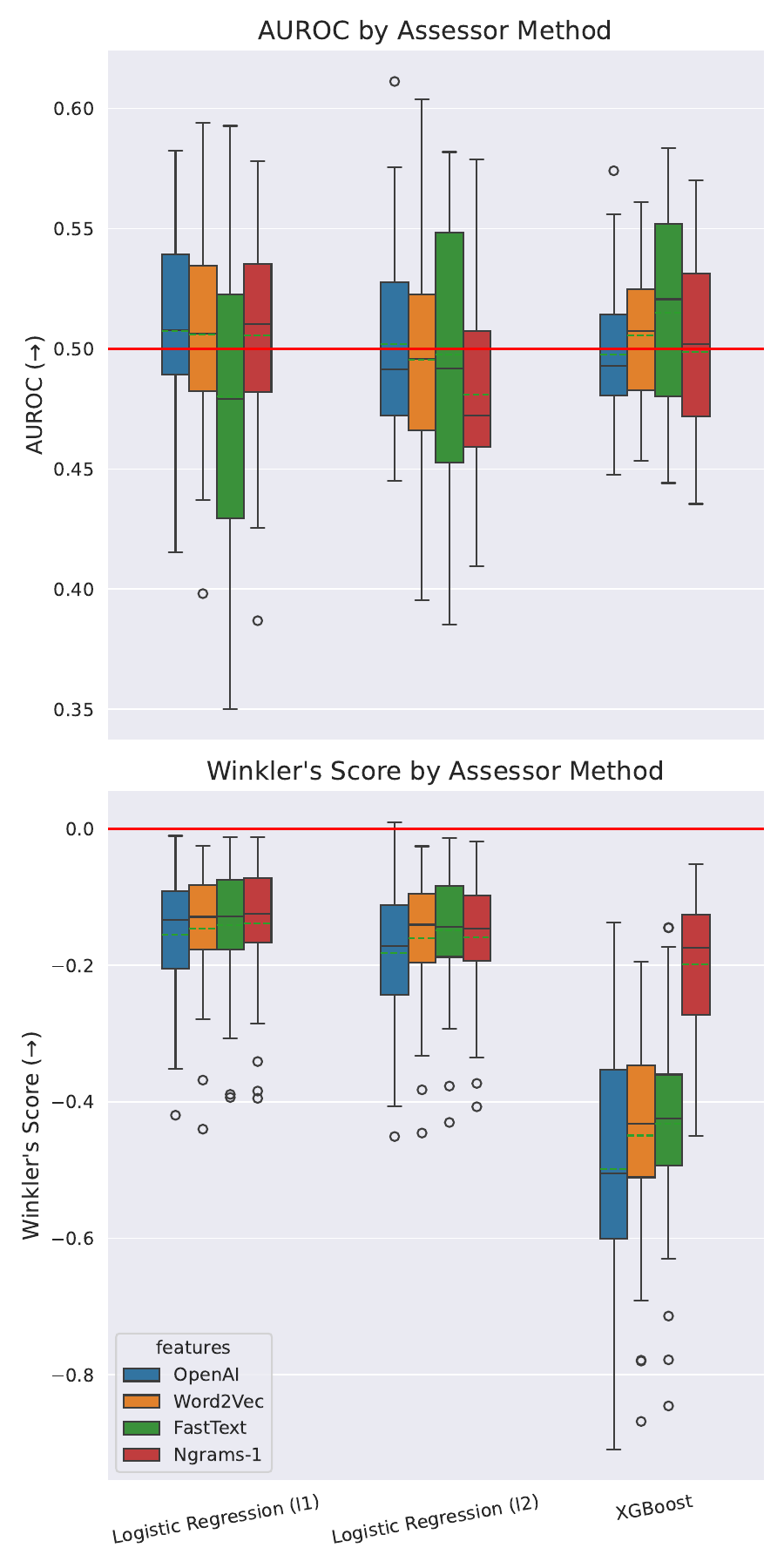}
    \caption{Distribution (over LLMs) of assessor-only performance metrics across different embedding schemes
    and classifiers, evaluated OOD on the BBH dataset.}
    \label{fig:experiments_AUROC_Brier_OOD}
\end{figure}

In this section we replicate the results of Figure \ref{fig:experiments_AUROC_Brier} for OOD assessment on the BBH dataset (see Figure~\ref{fig:experiments_AUROC_Brier_OOD}). 
The results are worse, as expected, especially in calibration. This simply encourages more work on further abstraction and different partitions when training the assessors, and \predbench is a tool for that. Note that OOD is a situation where simply extrapolating the accuracy of one benchmark to another is useless, and small improvements in OOD results can make a difference. 

\begin{table*}[tb]
    \centering
    \caption{List of LLMs used in our experimental setting. \predbench includes instance-level results for MMLU-Pro and BBH for all of them except the three GPT-4o versions, for which BBH results are unavailable. }
    \resizebox{0.82\textwidth}{!}{%
    \begin{tabular}{clccl}
        \toprule
        \textbf{Index} & \textbf{HuggingFace's Model Name} & \textbf{Family} & \textbf{Size} & \textbf{Version}\\
        \midrule
    1 & \texttt{calme-2.1-qwen2.5-72b} & Qwen & 72B & 2.5-Instruct \\
    2 & \texttt{DCLM-7B} & DCLM & 7B & \\
    3 & \texttt{gemma-2-2b-ORPO-jpn-it-abliterated-18-gemma-2-2b} & Gemma & 2B & v2 ORPO \\
    4 & \texttt{GPT-4o-2024-05-13} & GPT & - & 4o-2024-05-13 \\
    5 & \texttt{GPT-4o-2024-08-06} & GPT & - & 4o-2024-08-06 \\
    6 & \texttt{GPT-4o-mini} & GPT & - &  4o-mini \\
    7 & \texttt{GutenLaserPi} & Mistral & 7B & - \\
    8 & \texttt{Hermes-3-Llama-3.1-70B} & Llama & 70B & 3.1 \\
    9 & \texttt{Jallabi-34B} & Llama & 34B & v1.6 \\
    10 & \texttt{L3.1-ClaudeMaid-4x8B} & Llama & 4x8B & ClaudeMaid v1.0\\
    11 & \texttt{LayleleFlamPi} & Mistral & 7B & - \\
    12 & \texttt{leniachat-gemma-2b-v0} & Gemma & 2B & leniachat v0 \\
    13 & \texttt{leniachat-qwen2-1.5B-v0} & Qwen & 1.5B & leniachat v0 \\
    14 & \texttt{llama-3.1-8B-Galore-openassistant-guanaco} & Llama & 8B & 3.1-Galore-openassistant-guanaco \\
    15 & \texttt{Llama-3.1-8B-Lexi-Uncensored} & Llama & 8B & 3.1-Lexi-Uncensored \\
    16 & \texttt{Llama-3.1-8B-Lexi-Uncensored-V2} & Llama & 8B & 3.1-Lexi-Uncensored-V2 \\
    17 & \texttt{Llama-3.1-8B-MagPie-Ultra} & Llama & 8B & 3.1-MagPie-Ultra \\
    18 & \texttt{Llama-3-70B-Synthia-v3.5} & Llama & 70B & Synthia-v3.5 \\
    19 & \texttt{Llama-3-8B-Synthia-v3.5} & Llama & 8B & Synthia-v3.5 \\
    20 & \texttt{llama-3-luminous-merged} & Llama & 8B & luminous-merged \\
    21 & \texttt{Meta-Llama-3.1-70B-Instruct} & Llama & 70B & v3.1-Instruct \\
    22 & \texttt{Mistral-7B-v0.1-signtensors-1-over-4} & Mistral & 7B & v0.1 \\
    23 & \texttt{Mistral-7B-v0.1-signtensors-3-over-8} & Mistral & 7B & v0.1 \\
    24 & \texttt{Mistral-7B-v0.1-signtensors-5-over-16} & Mistral & 7B & v0.1 \\
    25 & \texttt{Mistral-7B-v0.1-signtensors-7-over-16} & Mistral & 7B & v0.1 \\
    26 & \texttt{neural-chat-7b-v3-1} & Neural Chat & 7B & v3-1 \\
    27 & \texttt{neural-chat-7b-v3-2} & Neural Chat & 7B & v3-2 \\
    28 & \texttt{neural-chat-7b-v3-3} & Neural Chat & 7B & v3-3 \\
    29 & \texttt{pythia-410m-roberta-lr\_8e7-kl\_01-steps\_12000-rlhf-model} & Pythia & 410M & -\\
    30 & \texttt{Qwen2.5-95B-Instruct} & Qwen & 95B & 2.5-Instruct \\
    31 & \texttt{qwent-7b} & Qwen & 7b & - \\
    32 & \texttt{solar-pro-preview-instruct} & Solar Pro & 22B & instruct \\
    33 & \texttt{SuperHeart} & Llama & 8B & v3.1 SuperNova-Lite \\
    34 & \texttt{Tess-3-7B-SFT} & Mistral & 7B & 3-SFT \\
    35 & \texttt{Tess-3-Mistral-Nemo-12B} & Mistral & 12B & Tess v3 \\
    36 & \texttt{Tess-v2.5.2-Qwen2-72B} & Qwen & 72B & Tess v2.5.2 \\
    37 & \texttt{Tess-v2.5-Phi-3-medium-128k-14B} & Phi & 14B & Tess v2.5.2.5 \\
    38 & \texttt{Trinity-2-Codestral-22B} & Mistral & 22B & Codestral v0.1 \\
    39 & \texttt{Trinity-2-Codestral-22B-v0.2} & Mistral & 22B & Codestral v0.1 \\
    40 & \texttt{VisFlamCat} & Mistral & 7B & - \\
    41 & \texttt{Yarn-Llama-2-13b-128k} & Llama & 13B & v2\\
        \bottomrule
    \end{tabular}
    }
    \label{tab:llms}
\end{table*}

\section{Failure analysis}
\label{app:failure}

Models that always fail or always succeed are highly predictable. This is why we use the AUROC and Winkler's score as assessor metrics, because they are balanced, useful to counteract this effect and compare assessors for models of different accuracy. However, can we still find that more performant models are more predictable, even after controlling for this?

\begin{figure*}[tb]
    \centering
    \includegraphics[width=0.7\linewidth]{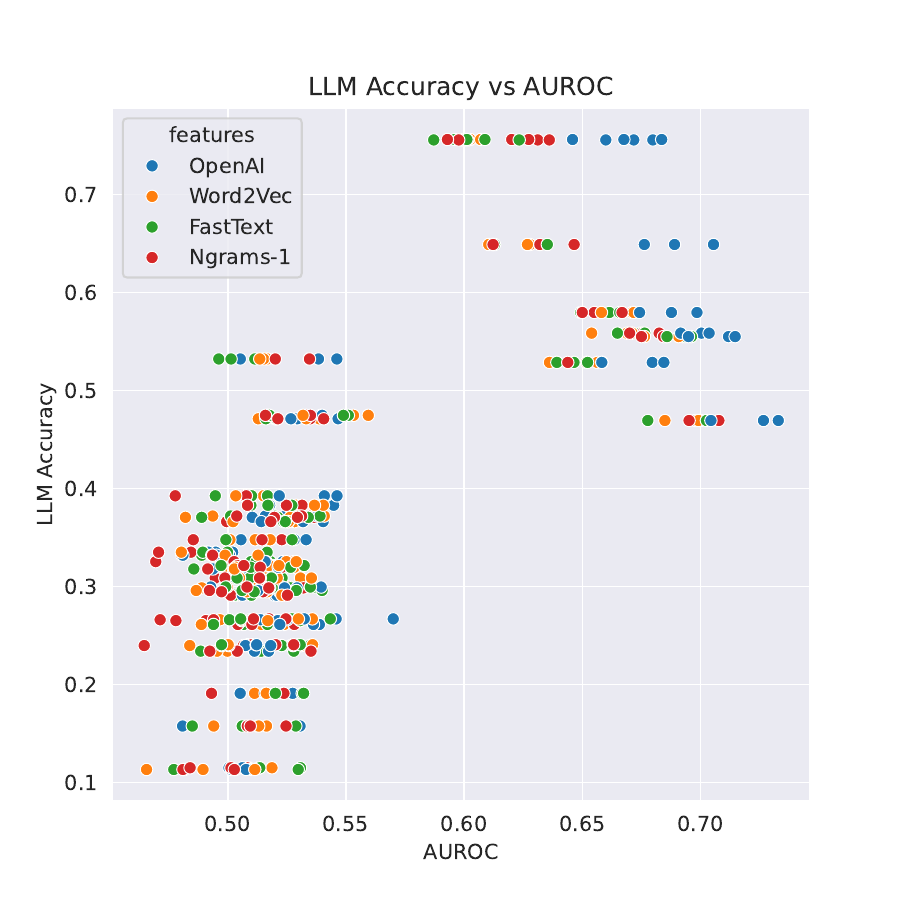}
    \caption{Relation between AUROC per assessor and accuracy per model on the test split of the MMLU-Pro dataset, for assessoras trained on the train split.}
    \label{fig:AUROC_accuracy}
\end{figure*}

We explore this question in Figure \ref{fig:AUROC_accuracy}, showing the relation between the accuracy and AUROC for all models and assessors respectively using the MMLU-Pro dataset (trained on the train split and evaluated on the test split). Recall that a classifier randomly guessing would produce AUROC of 0.5, while AUROC of 1 corresponds to perfect discrimination. We see a positive correlation, but the oriented interpretation is more interesting: assessors with high AUROC always correspond with models of high accuracy (the opposite is less clear). There are also two clear clusters in the plot, and the one on the top right has negative correlation. Moreover, in that cluster, the assessors using the most powerful features (the OpenAI embeddings) perform better.

\begin{figure*}[tb]
    \centering
    \includegraphics[width=0.7\linewidth]{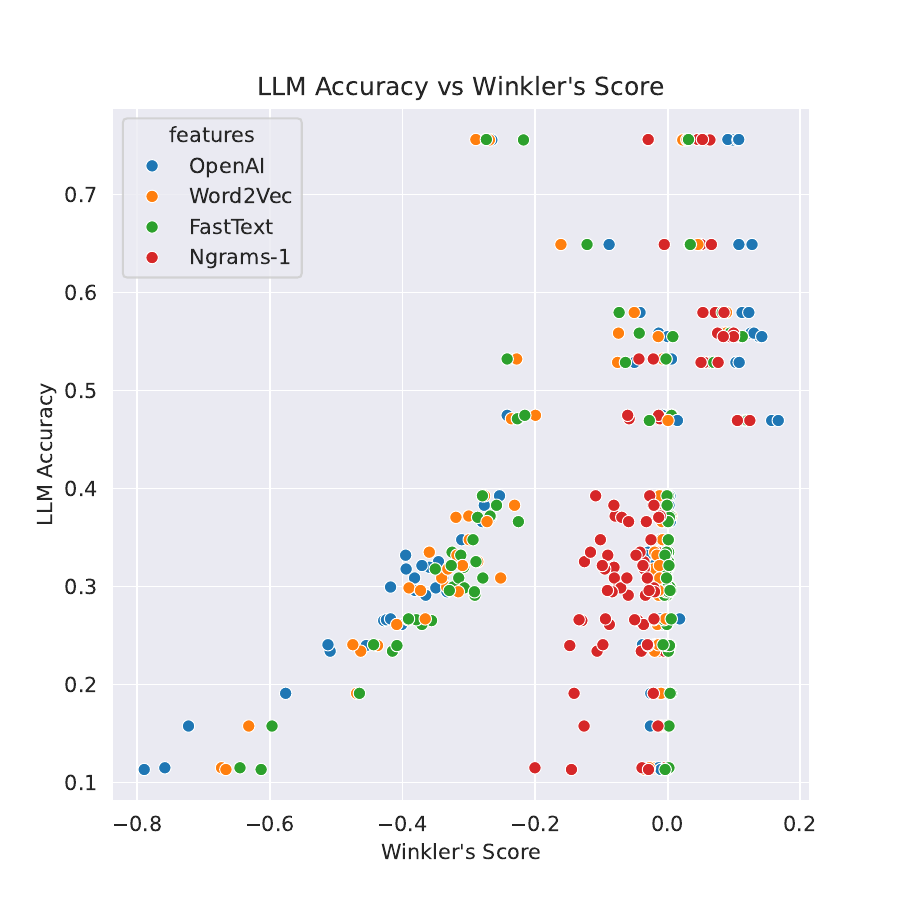}
    \caption{Relation between Winkler's score per assessor and accuracy per model on the test split of the MMLU-Pro dataset, for assessoras trained on the train split.}
    \label{fig:Winkler_accuracy}
\end{figure*}

Figure \ref{fig:Winkler_accuracy} replaces the AUROC with Winkler's score and shows a similar, but less clear behaviour, where again, higher Winkler's score implies higher LLM accuracy. Recall that, for the Winkler's score, a value of 0 corresponds to a constant assessor that always outputs the average accuracy; lower is worse and Winkler's score is 1 for perfect predictions. From this graph, two considerations can be made: 
\begin{enumerate}
    \item For the models with low accuracy, assessors are unable to get above the 0 (baseline level for a constant assessor) in terms of Winkler's score. Moreover, considering individually each LLM with low accruacy, the least powerful features (Ngrams-1) always lead to higher Winkler's score for the assessors, while the most powerful ones lead to much lower score (there is instead a mix for the two intermediate features). This suggests that most powerful features lead to the assessors overfitting on the training set, hence suggesting that these models have intrinsically poorly predictably.
    \item Instead, for the models with high accuracy, the opposite pattern can be observed: the most powerful features (the OpenAI embeddings) lead to higher Winkler's score, indicating that these features are encoding a general pattern impacting LLM performance.
\end{enumerate}

That influence suggests that we could inspect specific examples to see extreme differences between the actual and the predicted outcome. Indeed, assessors can be a very useful tool to analyse errors, such as using counterfactuals as in explainable AI (e.g., would the model fail if we changed the prompt in some particular way?). Also we can compare the confidence given by the assessor to different failures or successes. For instance, we can easily conduct a ranking of examples per benchmark according to their score predictability over different models and assessors, and how this relates to their difficulty (percentage of models that fail on the example). Theoretically, this should be related to the variance of a Bernoulli distribution, but it can be very insightful to explore deviations from this: failures of the model with high-confidence of success from the assessor and successes of the model with high-confidence of failure from the assessor.

To observe this, we selected the best LLM-assessor pair based on 0.8 PVR (\texttt{OpenAI/GPT-4o-2024-08-06} with Logistic Regression (l2) assessor based on OpenAI embeddings, see Fig.~\ref{fig:experiments_PPR}), and printed the lowest 5 instances in terms of assessor confidence which the LLM got right (Figure \ref{fig:lowest5correct}), and the highest 5 which the LLM got wrong (Figure \ref{fig:highest5wrong}). The high-confidence-but-wrong instances involve short questions, many of which seem straightforward, from several disciplines. In contrast, the low-confidence-but-correct instances involve very long question and many of them are law- or engineering-related. This shows how these failures, at least for this model, were ultimately caused by the base model and not by the assessor giving unreasonable estimates.

\begin{figure*}[tb]
    \centering
    \begin{tiny}
    \begin{Verbatim}[breaklines=true]
PROMPT 1400, Prediction: 0.35163472465364426:
Federal law prohibits "willingly and knowingly" taking cash in excess of $10,000 from the U.S. into a foreign country without first reporting the transaction in detail. An owner of a Detroit drug store takes his gross cash receipts each week into a city in Canada where he lives and does his banking. The office of the Deputy Atty. General learned that the owner was doing this, and indicted him on 10 counts of "willingly and knowingly" taking cash over $10,000 into a foreign country without reporting it. The owner's main defense is that he did not know of the law or that he was breaking it. The trial judge instructed the jury that mistake of law is no defense. He was convicted and appealed. Will the federal appellate court likely reverse the conviction?
A. No, because the owner's habitual actions imply intent to avoid reporting the cash.
B. No, the practice is so dangerous to the public interest that knowledge and specific intent are not required.
C. Yes, because willfulness clause requires proof of both knowledge of the law and a specific intent to commit the crime.
D. No, willfulness and knowledge are inferred by the habitual practice of transporting the cash.
E. Yes, because the owner was not intentionally breaking the law, he was simply unaware of it.
F. No, because ignorance of the law is not a valid defense.
G. Yes, because the owner is not a resident of the U.S. and therefore not subject to its laws.
H. No, because the owner is a business operator and therefore should be aware of such laws.
I. Yes, because treaties with Canada make all such reporting laws unenforceable.
J. Yes, because the owner was not given a fair chance to defend himself in court.
Answer:

PROMPT 1758, Prediction: 0.3642458853764882:
A man who owned a business believed that one of his employees was stealing computer equipment from the business. He decided to break into the employee's house one night, when he knew that the employee and her family would be away, to try to find and retrieve the equipment. The man had brought a picklock to open the employee's back door, but when he tried the door, he found that it was unlocked, so he entered. As the man was looking around the house, he heard sounds outside and became afraid. He left the house but was arrested by police on neighborhood patrol. What is the man's strongest defense to a burglary charge?
A. The back door to the house was unlocked.
B. The man was scared and left the house before committing a crime.
C. The man did not actually use the picklock.
D. The man was arrested outside, not inside, the house.
E. The man was only trying to retrieve his own property.
F. The man did not intend to commit a crime inside the house.
G. The man believed the stolen property was his.
H. The house was not occupied at the time of his entry.
I. The man did not take anything from the house.
Answer:

PROMPT 996, Prediction: 0.36962361800041293:
A wife is the beneficiary of a policy issued by an insurance company, insuring the life of her husband, now deceased. The policy contained a clause providing that double indemnity is payable in the event that death of the insured "results directly, and independently of all other causes, from bodily injury effected solely through external violent and unexpected means. "The husband was found dead in the chicken shed of his farm. His death resulted from wounds caused by a shotgun blast. The wife filed the necessary papers with the insurance company concerning proof of her husband's death. The insurance company admitted liability for the face amount of the policy but rejected the wife's claim for double indemnity. The wife then instituted suit against the insurance company demanding judgment according to the double indemnity provisions of the husband's insurance policy. At trial, the wife was called to testify about the events on the day of her husband's death. The wife said that she was in the kitchen when she heard a gunshot in the shed. As she rushed out of the house, she saw their neighbor running from the shed. The neighbor is present in court. As a witness, the wife was
A. competent, because she can provide a first-hand account of the incident.
B. incompetent, because she was not an eyewitness to the actual event.
C. incompetent, because her testimony is based on her perception of events.
D. competent, because she was present on the scene after the event occurred.
E. competent, because she had personal knowledge of the matter.
F. competent, because the neighbor is available to testify.
G. incompetent, because her testimony could potentially be biased.
H. incompetent, because she was testifying to facts occurring after her husband's death.
I. competent, because she can corroborate her account with the neighbor's testimony.
J. incompetent, because she had a personal interest in the outcome of the lawsuit.
Answer:

PROMPT 2162, Prediction: 0.38607054185111:
There is a state statute making it a misdemeanor "to falsely report a fire either intentionally or recklessly. " There were three college roommates who lived together in a small apartment. Two of the roommates decided to play a practical joke on the other roommate, which they liked to do from time to time because he was gullible. The two roommates were seated in the living room of their apartment. The other roommate was in an adjoining room and within earshot of the two roommates. Knowing that their roommate could hear their conversation, the two roommates falsely stated that a fire had been set at the student center at the college. After overhearing this conversation, the other roommate phoned the fire department and reported this information. Several fire trucks were dispatched to the college and determined the information to be false. If the two roommates are prosecuted for violating the aforementioned statute, they should be found
A. guilty, because they intentionally misled their roommate.
B. guilty, because they deliberately provided false information.
C. not guilty, because they did not directly call the fire department.
D. guilty, because they caused the false report to be made.
E. guilty, because they knowingly spread false information within earshot of their roommate.
F. guilty, because they are accomplices to their roommate.
G. not guilty, because it was a practical joke and they did not intend harm.
H. not guilty, because they didn't knowingly believe that their roommate would report the information to the fire department.
I. not guilty, because they didn't report the information to the fire department themselves.
J. not guilty, because they did not confirm the presence of a fire themselves.
Answer:

PROMPT 485, Prediction: 0.38794118768882585:
A defendant was on the first day of her new secretarial job when her boss called her into his office. The boss directly suggested that if the defendant did not go out on a date with him, she would be fired in one week. Every day during the remainder of the week, the boss approached the defendant with his demand, and the defendant refused to cooperate. At the end of the week, when the boss called the defendant into his office and again tried to pressure her to go out on a date with him, the defendant knocked him unconscious with a giant stapler and choked him to death. The defendant is tried for murder. In accordance with the following statute, the state relies at trial on the presumption of malice:"When the act of killing another is proved, malice aforethought shall be presumed, and the burden shall rest upon the party who committed the killing to show that malice did not exist. "If the defendant is convicted of first-degree murder and challenges her conviction on the grounds of the above statute, on appeal she will
A. lose, because the presumption may be rebutted.
B. win, because the statute violates due process.
C. lose, because the presumption of malice aforethought is constitutional.
D. win, because she acted in self-defense.
E. lose, because her actions were premeditated.
F. win, because the statute is unjust.
G. lose, because she did not show that malice did not exist.
H. win, because the statute is discriminatory.
I. lose, because she failed to overcome the presumption.
Answer:
\end{Verbatim}
    \end{tiny}
    \vspace{-0.7cm}
    \caption{Lowest 5 instances in terms of assessor confidence which the LLM got right. This is shown for the best LLM-assessor pair based on 0.8 PVR (\texttt{OpenAI/GPT-4o-2024-08-06} with Logistic Regression (l2) assessor based on OpenAI embeddings) and the MMLU-Pro dataset.}
    \label{fig:lowest5correct}
\end{figure*}

\begin{figure*}[tb]
    \centering
    \begin{tiny}
\begin{Verbatim}[breaklines=true]
PROMPT 365, Prediction: 0.9677333347366117:
A store sells two items for $10 each. One item costs $5.25, while the other costs $6.50. What ratio of items at each price must be purchased in order to have an average markup based on the selling price of 40%
A. 3 to 1
B. 4 to 3
C. 1 to 2
D. 3 to 2
E. 2 to 3
F. 2 to 5
G. 1 to 4
H. 1 to 3
I. 4 to 1
J. 5 to 3
Answer:

PROMPT 929, Prediction: 0.9578498958041702:
What is criterion related validity?
A. Criterion-related validity evaluates the test's ability to predict future or past performance.
B. Criterion-related validity is the testing of an individual's knowledge
C. Criterion-related validity measures the extent to which test scores are unaffected by external factors.
D. Criterion-related validity measures the test's consistency
E. Criterion-related validity assesses the degree to which a test captures a comprehensive range of abilities within a domain.
F. Criterion-related validity is the extent to which a test measures a theoretical construct or trait.
G. Criterion-related validity refers to the bias in a test's results
H. Criterion-related validity is the degree to which a test aligns with another widely accepted standard test.
I. Criterion-related validity is concerned with the extent to which a test correlates with a concurrent benchmark.
J. Criterion-related validity refers to the effectiveness of a test in predicting an individual's behavior in specified situations.
Answer:

PROMPT 1241, Prediction: 0.9559929993297027:
A store has 3 boxes of shirts. Each box has 4 packages with 7 shirts in each package. The expression 3 × (4 x 7) can be used to find the total number of shirts. Which expression can also be used to find the total number of shirts?
A. 14 × 3
B. 21 × 4
C. 12 × 7
D. 7 × 12
E. 14 × 4
F. 4 × 21
G. 12 × 3
H. 28 × 4
I. 28 × 7
J. 21 × 3
Answer:

PROMPT 902, Prediction: 0.9389530358257672:
Even though there is no such thing as a "typical cell" - for there are too many diverse kinds of cells - biologists have determined that there are two basic cell types. What are these two types of cells?
A. Single-celled and Multi-celled
B. Animal and Plant cells
C. Procaryotic and Eucaryotic
D. Diploid and Haploid cells
E. Photosynthetic and Non-photosynthetic cells
F. Vascular and Non-vascular cells
G. Prokaryotic and Eukaryotic
H. Somatic and Germ cells
I. Autotrophic and Heterotrophic cells
J. Aerobic and Anaerobic cells
Answer:

PROMPT 1229, Prediction: 0.9381125294629535:
Which of the following about meiosis is NOT true?
A. Meiosis produces two haploid gametes.
B. Homologous chromosomes join during synapsis.
C. Sister chromatids separate during meiosis I.
D. Crossing-over increases genetic variation in gametes.
Answer:


\end{Verbatim}
    \end{tiny}
    \vspace{-0.5cm}
    \caption{Highest 5 instances in terms of assessor confidence which the LLM got wrong. This is shown for the best LLM-assessor pair based on 0.8 PVR (\texttt{OpenAI/GPT-4o-2024-08-06} with Logistic Regression (l2) assessor based on OpenAI embeddings) and the MMLU-Pro dataset.}
    \label
{fig:highest5wrong}
\end{figure*}

\end{document}